\documentclass{article}

\usepackage{arxiv}

\usepackage[utf8]{inputenc} 
\usepackage[T1]{fontenc}    
\usepackage{hyperref}       
\usepackage{url}            
\usepackage{booktabs}       
\usepackage{amsfonts}       
\usepackage{nicefrac}       
\usepackage{microtype}      
\usepackage{lipsum}
\usepackage{graphicx}
\usepackage{amsmath}
\usepackage{algorithm}
\usepackage[noend]{algpseudocode}
\makeatletter
\def\BState{\State\hskip-\ALG@thistlm}
\makeatother
\graphicspath{ {./images/} }

\title{Modeling Survival in model-based Reinforcement Learning}

\author{
 Saeed Moazami \\
  Computer Science Department\\
  Lamar University\\
  \texttt{smoazami@lamar.edu} \\
   \And
 Peggy Doerschuk \\
  Computer Science Department\\
  Lamar University\\
  \texttt{pdoerschuk@lamar.edu} \\
}

\begin{document}
\maketitle
\begin{abstract}
Although recent model-free reinforcement learning algorithms have been shown to be capable of mastering complicated decision-making tasks, the sample complexity of these methods has remained a hurdle to utilizing them in many real-world applications. In this regard, model-based reinforcement learning proposes some remedies. Yet, inherently, model-based methods are more computationally expensive and susceptible to sub-optimality. One reason is that model-generated data are always less accurate than real data, and this often leads to inaccurate transition and reward function models. With the aim to mitigate this problem, this work presents the notion of survival by discussing cases in which the agent’s goal is to survive and its analogy to maximizing the expected rewards. To that end, a substitute model for the reward function approximator is introduced that learns to avoid terminal states rather than to maximize accumulated rewards from safe states. Focusing on terminal states, as a small fraction of state-space, reduces the training effort drastically. Next, a model-based reinforcement learning method is proposed (Survive) to train an agent to avoid dangerous states through a safety map model built upon temporal credit assignment in the vicinity of terminal states. Finally, the performance of the presented algorithm is investigated, along with a comparison between the proposed and current methods. 
\end{abstract}


\section{Introduction}
Owing to the recent advances in the development of algorithms and increasing computation power, we are witnessing a renaissance for reinforcement learning (RL) \cite{henderson2018deep}. These advances have made reinforcement learning algorithms capable of solving a wide range of complex decision-making problems in a broad area of domains such as robotics \cite{Kober}, game development \cite{Mnihatari,Silver}, finance \cite{Fischer}, healthcare \cite{Yu}, improving other learning methods \cite{elthakeb2018releq}, intelligent transportation \cite{Ault}, and many more. However, many aspects of reinforcement learning are still evolving. Sample efficiency and stability, for example, are being improved significantly by utilizing novel methods presented every few months.
The standard framework in the context of reinforcement learning is to learn an optimal policy to select actions to maximize a cumulative reward for a task \cite{sutton2018reinforcement}. A common trend in developing many reinforcement learning methods is that the agent must interact with the environment thousands or even millions of times to learn an optimal policy. This is specifically problematic in model-free reinforcement learning methods with high-dimensional function approximators, which has limited the application of reinforcement learning to simulations, video games, and a few physical systems. Learning from a relatively small number of faults to behave correctly is essential to reduce the risk of physical implementations and decrease computation costs.
Several efforts have been made to mitigate this problem seeking to reduce the risk of failure to make algorithms applicable for real-world everyday tasks. Categorized as safe reinforcement learning, a class of methods tries to minimize the risk of taking potentially dangerous actions \cite{garcia2015comprehensive}. Furthermore, a wide range of model-based reinforcement learning methods improve sample efficiency by reducing the required number of actual trials and errors through model-generated data \cite{Gu,Buckman,Feinberg}.
With the aim to contribute to this area, this paper explores the notion of survival in reinforcement learning by investigating the effectiveness of safety maps in order to improve the sample efficiency in model-based reinforcement learning methods. Firstly, it is discussed why in some dynamic environments, the only goal of the agent is to survive or, analogously, survival leads to higher returns and vice versa. Then, the possibility of controlling these environments only by avoiding dangerous states is discussed along with the merits of this method in sample efficiency improvement and its limitations. Finally, experiments are carried out on a selected benchmark to evaluate the proposed algorithm.

\section{Related Works}
\label{sec:headings}
The model-free class of reinforcement learning algorithms is often divided into value-based, policy gradient, and deterministic policy gradient methods, where the latest can be considered the combination of the first two. To enumerate a few effective algorithms from each category, Deep Q-Learning (DQN) \cite{Mnihatari} and its variants \cite{Hausknecht,van2016deep,Silver} are value-based, i.e., rely on value function approximator. TRPO \cite{SchulmanTrpo}, PPO \cite{SchulmanPPO}, A3C \cite{MnihAsynchronous}, and SAC \cite{Haarnoja} are policy gradient methods. Also, DDPG \cite{Lillicrap} and TD3 \cite{Fujimoto} belong to deterministic policy gradient algorithms.
Model-free reinforcement learning in particular, has been superior in terms of stability and optimality. However, sample efficiency has remained an issue, making this class of methods impractical in many cases due to their slow learning rates. Conversely, model-based reinforcement learning methods have proven to be significantly more efficient \cite{Wang}. Most of the recent model-based reinforcement learning methods use artificial neural networks as predictive models for the environment’s transition and reward function approximators. In Dyna-Q methods \cite{SuttonDyna}, these models provide imaginary rollouts for the agent to learn with fewer actual interactions with the environment. Approximate model-assisted neural fitted Q-iteration (AMA-NFQ) \cite{Lampe} utilizes a learned model to generate imaginary trajectories to update a Q-function and shows significant improvement in CartPole \cite{BartoCartpole} benchmark sample efficiency.
Using these imaginary rollouts improves model-free reinforcement learning effectively if the model’s predictions are accurate, and deteriorates the performance when the model is inaccurate \cite{Janner,Gu}. To address this problem, the model-based value expansion method \cite{Buckman} controls the depth of imagination to improve the performance of the model-based method by keeping the model-generated data accurate enough. In continuous deep Q-learning with model-based acceleration \cite{Gu}, authors present normalized advantage functions (NAF) as a continuous Q-Learning method. They use iteratively refitted local linear learned models to accelerate the learning process. Nonetheless, the majority of model-based reinforcement learning algorithms rely mainly on training models to approximate reward function along with transition models. Taking one step forward, in this paper, we observe situations in which we can avoid training a reward function and substitute it with a safety map. Also, we use shallow depth predictive models for transitions to improve accuracy and prevent using predictive models for unnecessary states.

\section{Background}
The current reinforcement learning problem is defined as a Markov decision process (MDP) with a tuple \(\left \langle  S,A,p,r,\rho_0,U\right \rangle \), where S is the state-space and \(A\) is the discrete action space, and \(N=\left|A  \right|\) is the number of possible actions. \(p\) represents the state-transition probability density \(p:S\times S\times A\rightarrow\left[0,\infty\right)\) of the next state \(s_{t+1}\in S\), given the current state  \(s_t\in S\) and action \(a_t\in A\), \((s_t\longmapsto s_{t+1}|a_t)\). The agent receives a reward \(r\) while making a transition \(s_t\longmapsto s_{t+1}\). Accordingly, the reward function and the return will be defined as, \( r_t=r\left(s_t,a_t,s_{t+1}\right)\) and \(R\left(\tau\right)=\sum_{t=0}^{H}r\left(t\right)\) respectively. \(H\) is the horizon or the final time step, and trajectory \(\tau=\left(s_0,a_0,s_1,a_1,...\right)\) is a sequence of state actions in the environment where \(s_0\) is the initial state from the initial state distribution denoted by \(\rho_0\). Also, \(U\) is the set of terminal states in which an episode of state actions ends. The agent uses the policy \(\pi\) to decide what action to take based on the observed state, i.e., \(a_t=\pi\left(s_t\right)\). In this setup, we aim to find an optimal policy \(\pi^\ast\) through which the agent maximizes the expected return.\\
In value-based model-free reinforcement learning, algorithms calculate values for states by building a value function \(V^\pi\left(s\right)\), which gives the expected return if the agent starts from state \(s\) and takes actions according to policy \(\pi\). Similarly, a value function can be defined based on state-action pairs as \(Q^\pi\left(s,a\right)\). Subsequently, the algorithm aims to find an optimal policy \(\pi^\ast\) to select actions based on the calculated values in different states in order to maximize the expected return as:

\begin{equation}
    Q^\ast\left(s,a\right)=\underset{\pi}{\max} E \left [R\left(\tau \right)\left|s_0=s,a_0=a\right.\right],
\end{equation}

or

\begin{equation}
    V^\ast\left(s\right)=\underset{\pi}{\max} E \left[R\left(\tau\right)\left|s_0=s\right.\right].
\end{equation}

In this case, the optimal policy will be \(\pi^\ast\left(s\right)=\arg \underset{a}{\max} {Q^\ast}\left(s,a\right)=\pi^\ast\left(s\right)=\arg \underset{a}{\max} {V^\ast}\left(s\right)\).
Variants of value-based methods usually use an artificial neural network (ANN) as a function approximator parameterized by \(\theta\) to calculate \(Q_\theta^\pi\left(s,a\right)\). \(\theta_k\) is typically a set of weights for the ANN for iteration \(k\), which is iteratively calculated using the Bellman update:
\begin{equation}
Q_{\theta_{k+1}}^\pi\left(s_t,a_t\right)\gets Q_{\theta_k}^\pi\left(s_t,a_t\right)+\alpha\left[r\left(s_t,a_t,s_{t+1}\right)+\gamma\underset{a_{t+1}}{\max}{Q^\ast}\left(s_{t+1},a_{t+1}\right)-Q_{\theta_k}^\pi\left(s_t,a_t\right)\right].
\end{equation}

In policy optimization model-free methods, considering \(\pi_\phi\) as a policy, parametrized by \(\phi\), the goal is to maximize the expected return as \(J(\pi_{\phi}) = \underset{\tau \sim \pi_{\phi}}{\mathbb{E}} \left[ {R(\tau)} \right]\). A gradient ascent method is usually used to update policy parameters in the direction of the return’s increase as:

\begin{equation}
\phi_{k+1}=\phi_k+\alpha\left.\nabla_\phi J\left(\pi_\phi\right)\right|_{\phi_k}.
\end{equation}

The term policy gradient, \(g=\nabla_\phi J\left(\pi_\phi\right)\) is usually calculated by collecting several trajectories as \(\mathcal{T}=\tau_i,i=1,2,..,N\) to estimate a sample mean as:

\begin{equation}
    \hat{g} = \frac{1}{|\mathcal{T}|} \sum_{\tau \in \mathcal{T}} \sum_{t=0}^{H} \nabla_{\phi} \log \pi_{\phi}(a_t |s_t) R(\tau),
\end{equation}

where \(|\mathcal{T}|\) is the number of sampled trajectories in the environment using the policy \(\pi_\phi\).\\

Although there are several different model-based reinforcement learning methods in the literature, almost all of them rely on transition and reward function models. Transition model \(\hat{p}\left(s_t,a_t\right)\) predicts future states using current state and action, and reward function model \(\hat{r}\left(s_t,a_t\right)\) predicts the expected reward by taking action \(a_t\) in state \(s_t\). The majority of model-based methods use actual rollouts to train these models. Afterward, they use trained models to generate data instead of the environment to reduce the number of actual rollouts needed \cite{SuttonDyna}.

\section{METHODOLOGY}
\begin{figure}[H]
  \centering
  \includegraphics[height=1.5 in,keepaspectratio]{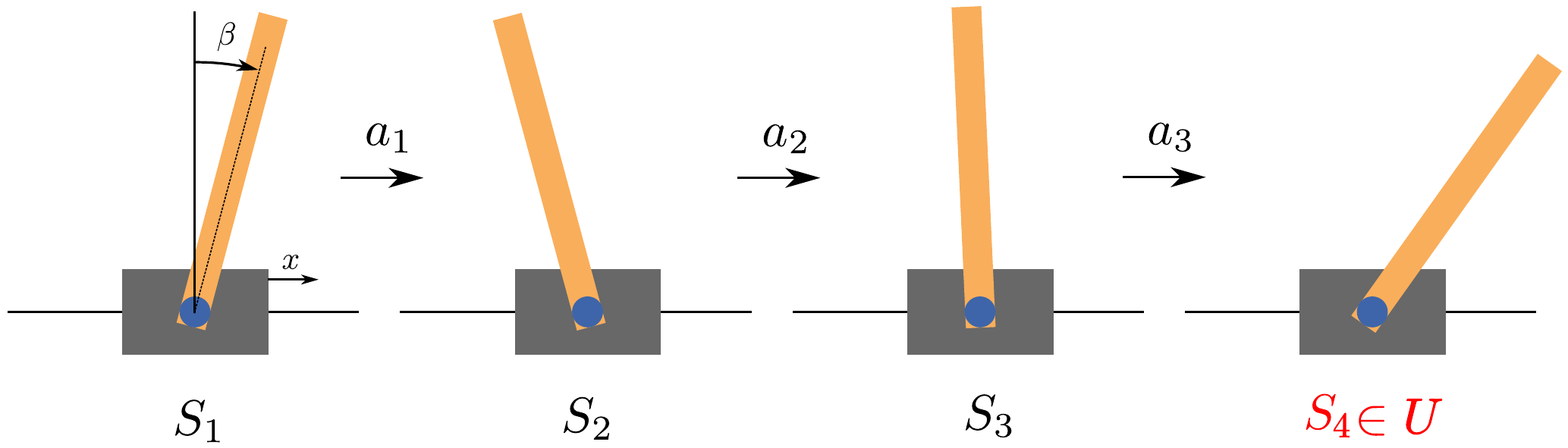} 
  \caption{The environment emits positive rewards through \(s_1\) to \(s_3\) transitions and a negative reward from \(s_3\) to \(s_4\in U\).}
  \label{fig:CartPole}
\end{figure}

While the objective of reinforcement learning agents is to maximize the return, in some environments, this is analogous to survival. There are many examples in the literature that match this intuition. The CartPole environment, which is widely used in the verification of reinforcement learning algorithms \cite{Gu,Behzadian} is an example. As it is shown in fig.\ref{fig:CartPole}, the CartPole has a four dimensional state space \((x, \dot{x},  \beta, \dot{\beta})\), where if the agent visits \(\left|\beta\right|\geq15\) or \(\left|x\right|\geq2.4\) the episode fails. Otherwise, if the agent stays alive longer by balancing the pole, it will receive more rewards.

\subsection{Reward function model efficiency}
Commonly, model-based reinforcement learning algorithms generate data required by model-free methods, and as model-free methods generally rely on a reward function, it is essential to have an accurate approximation for the reward in addition to the transition predictor. A side effect of this necessity is that a less precise model to predict reward can lead to poor performance. This problem can be more severe in unbalanced reinforcement learning when reward signals can be either negative or positive in relatively close states \cite{SuttonTemporal}. Hence, we can decouple different classes of reward signals to have a more efficient and accurate model. fig.\ref{fig:CartPole} shows how a decoupled reward can lead to faster and more precise reward approximation in the CartPole problem. The environment emits positive rewards through \(s_1to s_3\) transitions and a negative reward from \(s_3\) to \(s_4\in\ U\). Instead of training a model to predict both types of rewards, we can use only the negative rewards to train a model to predict if a state is dangerous or not.

\subsection{State-action sample efficiency}
\begin{figure}[H]
  \centering
  \includegraphics[height=2 in,keepaspectratio]{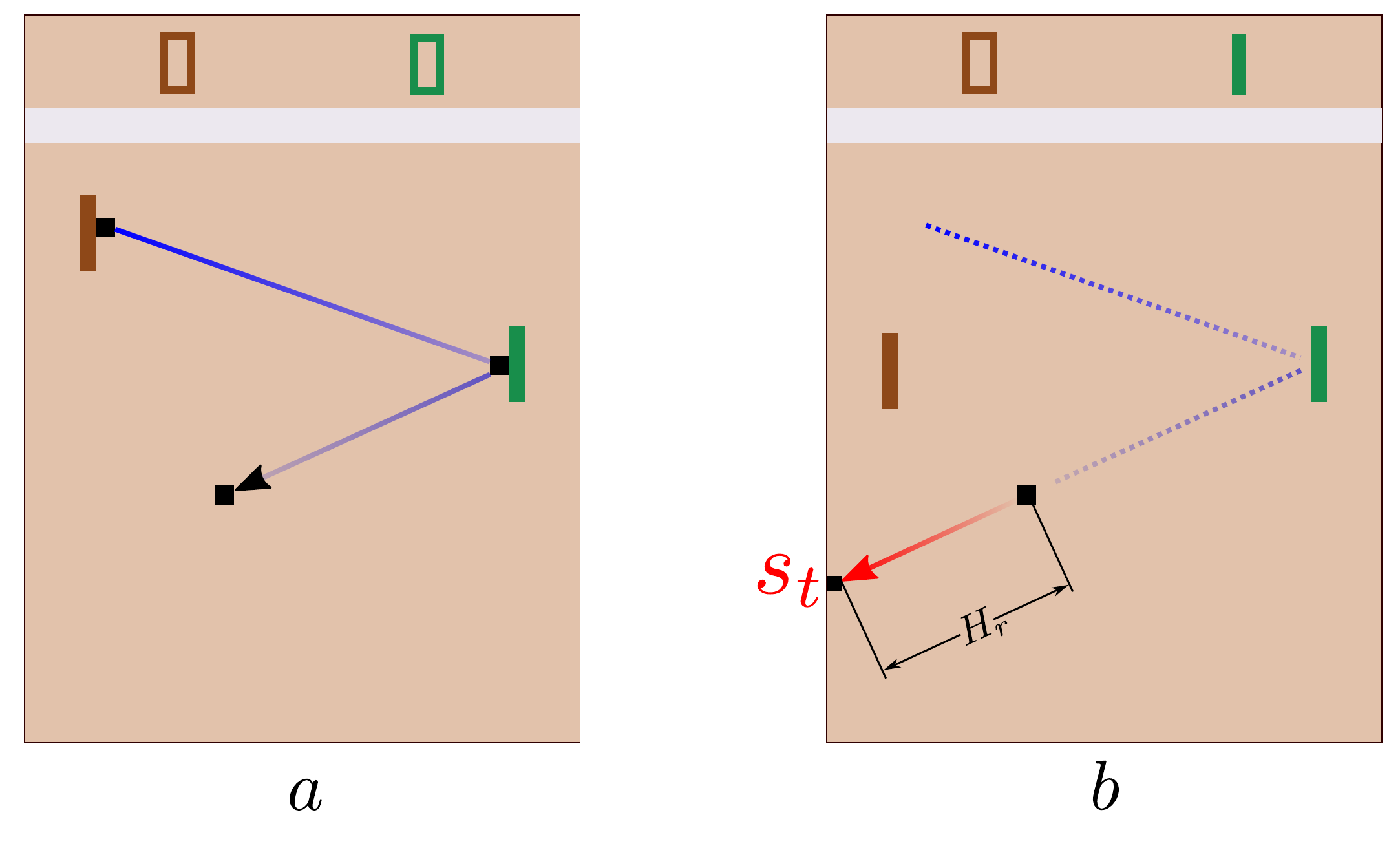}
  \caption{a: demonstrates a safe path for the ball with no valuable information for the agent to learn from.
b: a finite number of \((H_r)\) state actions that can lead to a negative reward or termination. If the agent learns how to survive in these states, consequently, it can maximize its reward.
}
  \label{fig:Pong}
\end{figure}

Consider the Atari Pong game illustrated in fig.\ref{fig:Pong} as another example. The agent must prevent the ball from passing its paddle by taking appropriate actions to place itself in front of the ball. As illustrated in fig.\ref{fig:Pong}.a, after hitting the ball back, there are several time steps in which the agent must not take any specific actions for higher rewards. While the ball goes across the field to reach the opponent and returns to the vicinity of the paddle, the agent can take any action without losing points. In fact, there are many situations in which there is nothing for the agent to perform. Consequently, these state-action pairs in this environment do not necessarily contribute to the training of a presumed optimal policy. Now imagine that the ball passes the agent’s paddle and reaches the state shown as \(u\in\ U\) in fig.\ref{fig:Pong}.b. The agent receives a negative reward, and one episode terminates. The agent needs data from multiple past time steps to have enough time to prevent this from happening. Therefore, using data in the vicinity of the terminal state would be required to train the agent to accomplish this task.

\subsection{Modeling survival}
Using the observations mentioned above, we discuss the notion of survival in discrete action space model-based reinforcement learning. To that end, we define a risk function, indicating how risky it is to take each action in any state. Assume that \(\mathcal{R}^\pi\left(s_t,a_t\right):S\times\ A\rightarrow\left(0,1\right]\) quantifies the risk of taking action \(a_t\) under the policy \(\pi\) while being at state \(s_t\) in time step \(t\). Intuitively, \(\mathcal{R}^\pi\) measures how close is the agent to a state-action pair with negative reward from the current state by taking actions based on the current policy. As we discussed, we assume that negative rewards are associated with taking risky actions in some states, and the environment omits negative rewards at transitions to terminal states. Thus, we can write \(\mathcal{R}^\pi\left(s_t,a_t\right)\) as:

\begin{equation}
    \mathcal{R}^\pi\left(s_t,a_t\right)=Pr\left(s_t\rightarrow u\middle|\ a_t \sim \pi\right),u\in\ U.
\end{equation}

\begin{figure}[H]
  \centering
  \includegraphics[height=2 in,keepaspectratio]{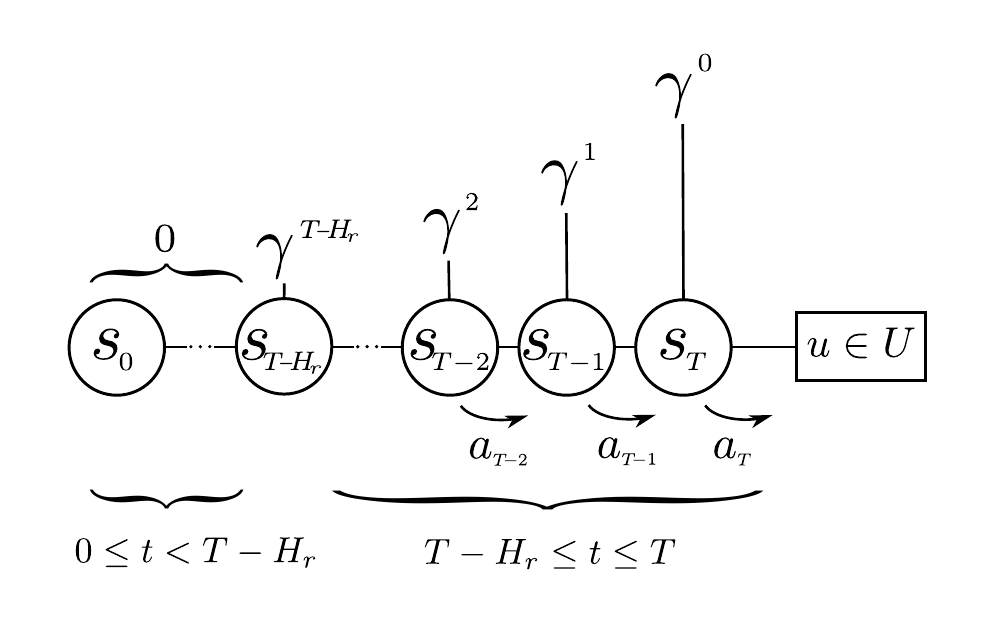}
  \caption{State action sequence of one episode and credit assignment to \(H_r\) last state-actions in a discounted manner.}
  \label{fig:Reverse_Horizon}
\end{figure}

In this case, we aim to train the agent in a way to learn how to avoid taking risky actions by avoiding termination. Let \(\mathcal{R}^\pi\left(s_t,a_t\right)=0\) and \(\mathcal{R}^\pi\left(s_t,a_t\right)=1\) indicate that it is entirely safe and extremely risky to take action \(a_t\) in the state  \(s_t\), respectively. Also, using an optimistic view, we assume that initially, all actions are safe if unless otherwise is indicated during experimenting in the environment; and \(\forall\ u\in\ U:\ \mathcal{R}^\pi\left(s_t,a_t\right)=1 \text{ if } \left(s_t\longmapsto u|.\right)\).
Similarly, a value can be assigned to indicate how dangerous is a state using \(D^\pi\left(s_t\right)\), which gives the danger of being in a particular state instead of the risk of taking specific action in a state. Then we consider a state dangerous if an arbitrary action has led to termination from that state.
Intuitively, a negative reward can be the result of actions that have been taken many timesteps before. The temporal credit assignment problem and eligibility traces are longlasting topics in reinforcement learning that discuss which actions contribute to a delayed reward signal \cite{SuttonTemporal,sutton2018reinforcement}. In this paper, it is assumed that reverse horizon, denoted as \(H_r\), is a finite number of action state pairs that contribute to a negative reward. As it is depicted in fig.\ref{fig:Reverse_Horizon}, the credit is assigned to the vicinity of negative rewards in reverse discounted manner, where \(0<\gamma<1\) is the discount factor.

Then, we calculate \(D^\pi\) for a sampled episode as the following: 

\begin{align}\label{eq7}
  D^\pi\left(s_t\right)=
  \begin{cases}
  \gamma^{T-t} & \text{if $T-H_r\le t \le T$}, \\
  0 & \text{if $0\le t<T-H_r$}
  \end{cases}.
 \end{align}

To avoid a foregone conclusion, after each rollout, we update a function approximator \(D^\pi\left(s_t\right)\) towards the calculated value as:

\begin{equation}
    D_{\psi_{k+1}}^\pi\left(s_t\right)\gets\ D_{\psi_k}^\pi\left(s_t\right)+\alpha(1-D_{\psi_k}^\pi\left(s_t\right)),
\end{equation}

where \(\psi_k\) is the parameter vector that defines \(D^\pi\) for \(k^{th}\) iteration, and \(0<\alpha\ll1\) is the learning rate. Having an approximation of how dangerous is being in a state, we use a trained transition model \(\hat{p}\) to predict the future states for all possible actions, then try to avoid dangerous states by taking the action that yields the least danger for the agent.

\subsection{Implementation}
\begin{figure}[H]
  \centering
  \includegraphics[height=2 in,keepaspectratio]{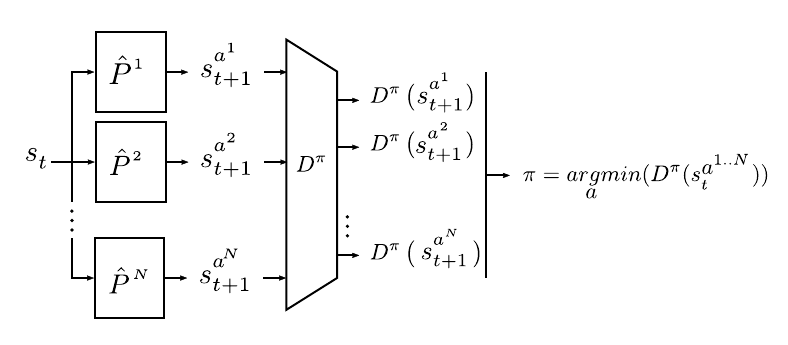}
  \caption{Neural network structure for predicting the future states based on the current state and different actions, and approximating the danger associated with each future predicted state.
}
  \label{fig:NN_Structure}
\end{figure}

A reply buffer memory \(\mathcal{M}\) is utilized to record sampled rollouts as \(\left \langle s_t,a_t,s_{t+1} \right  \rangle \). Then transition prediction models \({\hat{p}}^{1..N}\) are trained using randomly sampled \((s_t\longmapsto\ s_{t+1}|a_t)\) from \(\mathcal{M}\) to learn to predict \({\hat{s}}_{t+1}^{1..N}={\hat{p}}^{1..N}\left(s_t,a_t^{1..N}\right)\), where \(N=|A|\) is the number of possible actions. The discrete action space of the problem enables a separate neural network structure for each action. As shown in fig.\ref{fig:NN_Structure}, a single independent ANN learns to approximate \(D^\pi\) for all possible predicted future states, \({\hat{s}}_{t+1}^{1..N}\).

A secondary memory buffer \(\mu\) records \( \left \langle s_t,a_t,s_{t+1} \right \rangle\) of all timesteps of the last episode. To train \(D^\pi\), an equal number of \(H_r\) state action pairs are used from the vicinity of the reward based on Eq.\ref{eq7} and  \(H_r\) random samples from safe states, i.e., \(D^\pi=0\). The policy returns the action with the minimum calculated \(D^\pi\) among all predicted future states for possible actions, while an \(\epsilon\)-greedy policy ensures the exploration. Algorithm 1 represents the pseudo-code for the presented method.

\begin{algorithm}\label{alg}
\caption{Model-Based Survival}\label{euclid}
\begin{algorithmic}
\State {Initialize $D^\pi$ and $\hat{P}^{1..N}$ parameters, $\psi$, and $\phi^{1..N}$ }
\State {Replay buffer $\mathcal{M} \gets \emptyset$}

\For{$ each episode $} 
    \State $\textit{episode buffer } \mu \gets \emptyset$.
    \For{each time step $t$} 
        \State $\text{Future state danger set } \mathcal{S} \gets\emptyset $
        \For{$ i \text{ in } \left\{1,..,N\right\}$}
            \State $\hat{s}_{t+1}^i={\hat{P}}^i(s_t)$
            \State $\mathcal{S}=\mathcal{S}\cup\left\{ D({\hat{s}}_{t+1}^i) \right\}$
            \State $ a \gets \text{argmin} (\mathcal{S}) \text{( take the action that yields the least danger)}$
        \EndFor
        \State $s_{t+1}\gets \text{step(a) (take one step in the environment using action } a$
        \State $\mathcal{M}= \mathcal{M} \cup \left\{ \left \langle s_t,a_t,s_{t+1} \right \rangle \right \}$        
        \If {$s_{t+1} \in u: $}
            \State $head \gets$ last $H_r$ states from $\mu$
            \State $tail \gets$ Randomly sampled $H_r$ safe states (not from the head)
            \State {Train $D$ using Calculated value from:}   
                    \begin{align} \notag
                      D^\pi\left(s_t\right)=
                      \begin{cases}
                      \gamma^{T-t} & \text{if $T-H_r\le t \le T$ (head)} \\
                      0 & \text{if $0\le t<T-H_r$ (tail)}
                      \end{cases}.
                     \end{align}           
            \State {Train ${\hat{P}}^{1..N}$ using random samples of $\left \langle s_t,a_t,s_{t+1} \right \rangle$ from Replay buffer $\mathcal{M}$}
            \State{break (next episode).} 
            
        \EndIf
    \EndFor
\EndFor

\end{algorithmic}
\end{algorithm}

\section{Simulation Results}
\begin{figure}[H]
  \centering
  \includegraphics[height=2.2 in,keepaspectratio]{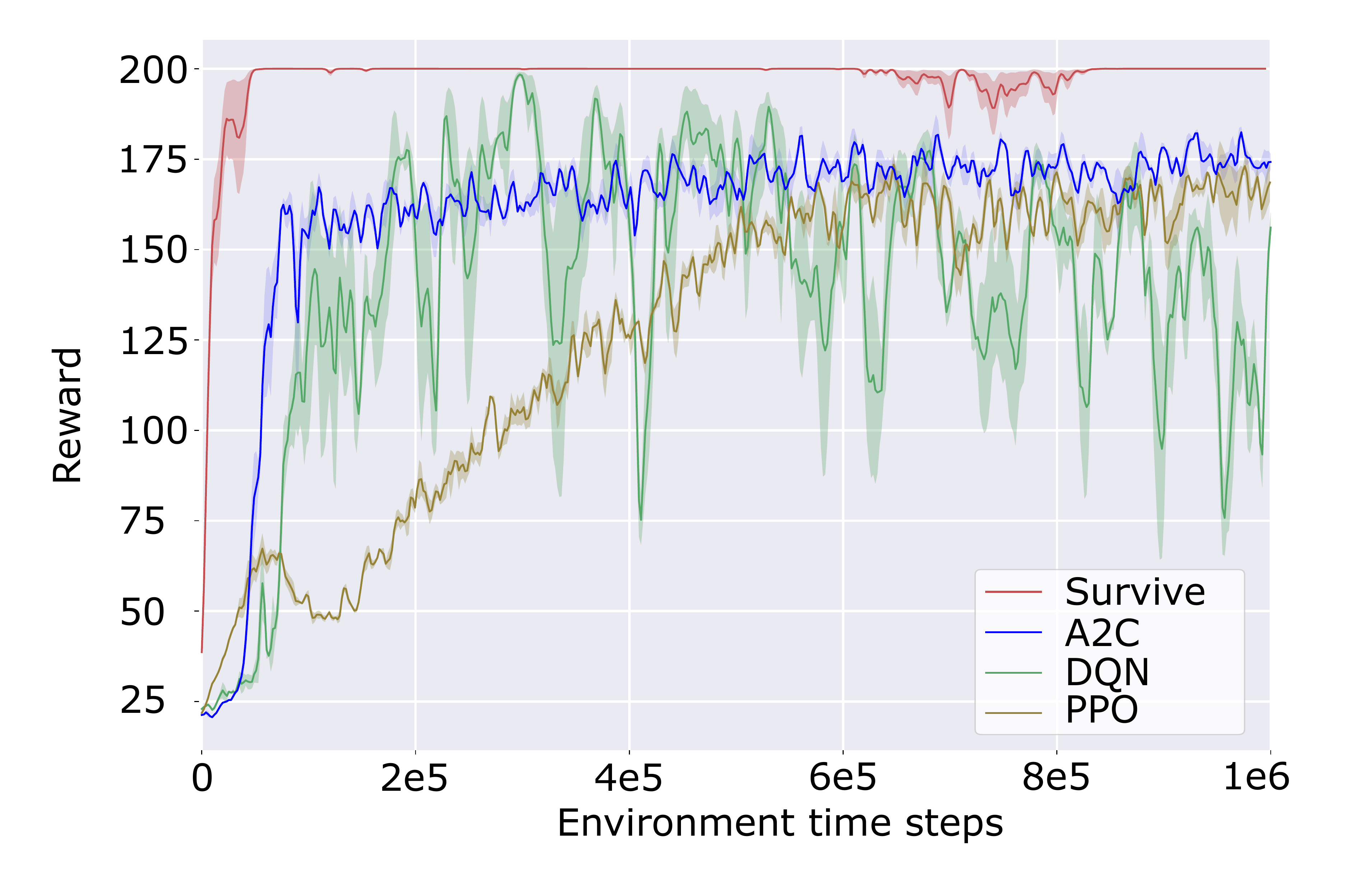}
  \caption{Learning rate comparison of Survive (Ours), TRPO, PPO, DQN algorithms in controlling OpenAI CartPole.}
  \label{fig:Performance}
\end{figure}

The OpenAI\(^{\textregistered}\) gym \cite{Brockman} CartPole environment has been used for the purpose of evaluating the presented method. Also, OpenAI baselines \cite{dhariwal2017openai} implementation of proximal policy optimization (PPO), deep Q-learning, and actor-critic (A2C) methods have been utilized to compare the performance of the algorithms. As it can be seen from fig.\ref{fig:Performance}, the presented method (Survive) shows higher performance in terms of sample efficiency compared to existing model-free methods in the literature in controlling the CartPole problem. The learning process takes place in a significantly lower number of time steps, and the agent achieves a higher and more stable reward average. 

\begin{figure}[H]
  \centering
  \includegraphics[height=3.5 in,keepaspectratio]{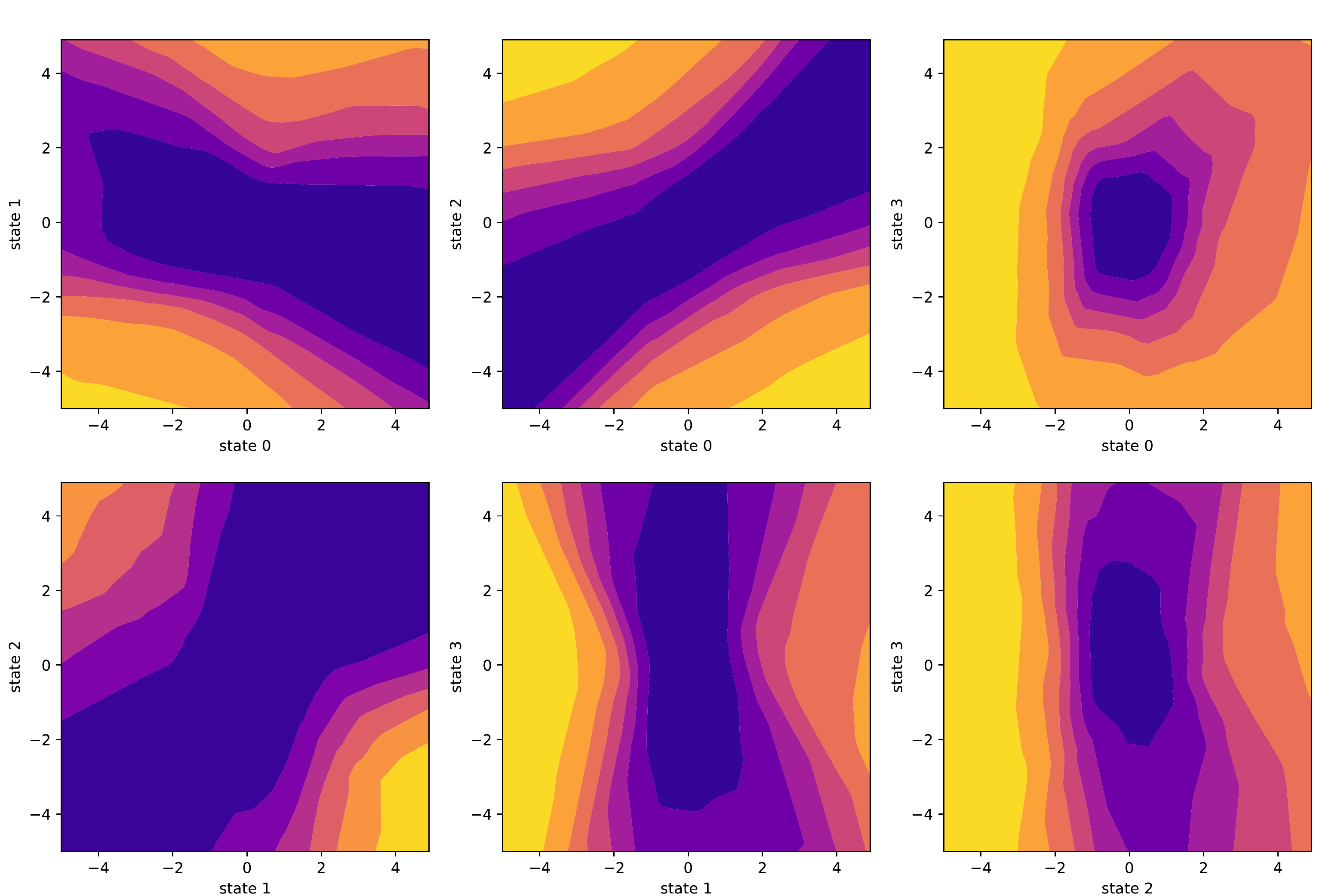}
  \caption{Danger map contours of the CartPole problem. The agent evaluates the future states and avoids lighter areas, i.e., states with higher \(D\) values.}
  \label{fig:2D_plot}
\end{figure}

Fig.\ref{fig:2D_plot} gives the intuition of danger maps by depicting the calculated \(D\) contours for all possible 2D combinations of the CartPole's 4-dimensional state space, where state 0,1,2, and 3 are the cart position, velocity, the pole angle, and angular velocity respectively. A general tendency to keep the states closer to zero is evident in the illustrations, indicating an optimal state for survival as the cart is positioned in the middle of the track, the pole is perfectly balanced, and neither the cart nor the pole is moving. Asymmetrical contours imply that the agent has been trained to successfully accomplish the task without observing all of the dangerous states. Also, it can be noted that the safe area is elongated in the direction of state 0, showing that the CartPole can survive in different positions.

\begin{figure}[H]
  \centering
  \includegraphics[height=3.2 in,keepaspectratio]{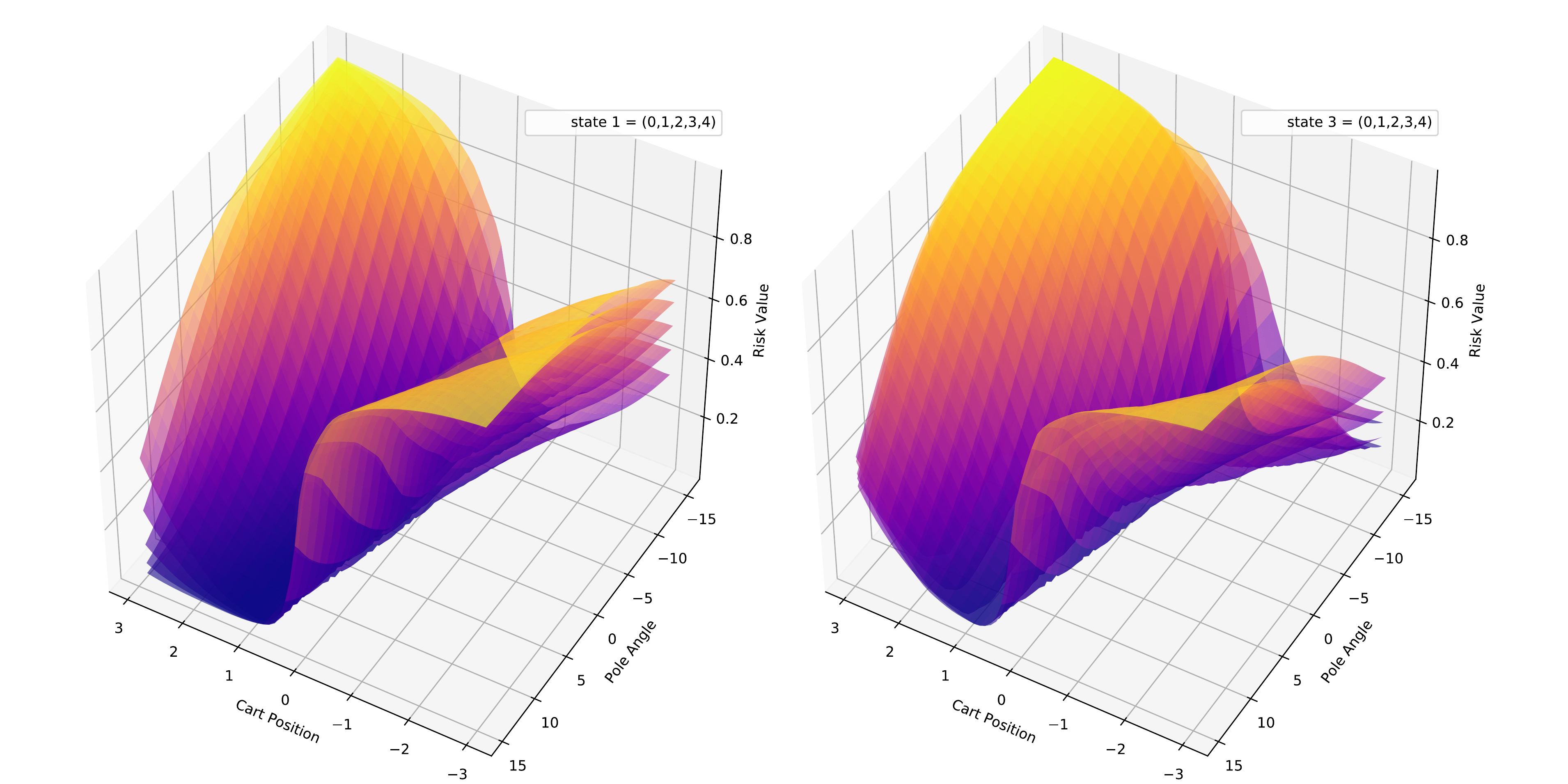}
  \caption{Calculated values for \(D\) for the cart position and pole angle in  the CartPole problem, for different values of left, cart velocity, and right, pole angular velocity.}
  \label{fig:3D_plot}
\end{figure}

In fig.\ref{fig:3D_plot}, danger maps are illustrated based on cart position and pole angle for different values of cart’s linear and pole’s angular velocities \( \left\{0,1,..,4 \right \} \), respectively . The algorithm presents the required robustness to keep the agent away from dangerous states, i.e., high values of cart position and pole angel which matches the background knowledge for this problem. Semi-flat areas of high risk values with lighter colors implicate that the agent has concluded that some states are extremely dangerous, $D \simeq 1$ independent from other states' value. 

\section{Conclusion and Future Work}
This paper discusses the notion of survival in model-based reinforcement learning by investigating situations in which the agent can succeed in an environment by avoiding termination. The idea has been implemented through sampling in the vicinity of terminal states by assigning risk values to the states that have led to termination. A model-based reinforcement learning algorithm has been used to train an agent to evaluate the future state for all possible actions and select the action that will lead to a state that yields the least danger.  Also, the model does not rely on a trained reward function, which makes it less prone to poor performance due to the reward model inaccuracy. The Algorithm has been applied to the CartPole problem and has successfully solved the problem. The Survive algorithm shows higher performance in sample efficiency, stability, and reward maximization compared to model-free methods.\\

The current setup has limited the application of this method to environments that match specific criteria, i.e., those that can win by staying alive. However, the idea can be propagated to a more comprehensive set of situations by integrating the notion of survival with reward maximization, and this paper provides the basis for generalizing this idea.

\bibliographystyle{unsrt}  
\bibliography{Survival}  

\begin{thebibliography}{10}

\bibitem{henderson2018deep}
Peter Henderson, Riashat Islam, Philip Bachman, Joelle Pineau, Doina Precup,
  and David Meger.
\newblock Deep reinforcement learning that matters.
\newblock In {\em Thirty-Second AAAI Conference on Artificial Intelligence},
  2018.

\bibitem{Kober}
Jens Kober, J~Andrew Bagnell, and Jan 
  Robotics~Research Peters.
\newblock Reinforcement learning in robotics: A survey.
\newblock 32(11):1238--1274, 2013.

\bibitem{Mnihatari}
Volodymyr Mnih, Koray Kavukcuoglu, David Silver, Andrei~A Rusu, Joel Veness,
  Marc~G Bellemare, Alex Graves, Martin Riedmiller, Andreas~K Fidjeland, and
  Georg 
\newblock Human-level control through deep reinforcement learning.
\newblock 518(7540):529--533, 2015.

\bibitem{Silver}
David Silver, Thomas Hubert, Julian Schrittwieser, Ioannis Antonoglou, Matthew
  Lai, Arthur Guez, Marc Lanctot, Laurent Sifre, Dharshan Kumaran, and Thore
\newblock A general reinforcement learning algorithm that masters chess, shogi,
  and go through self-play.
\newblock 362(6419):1140--1144, 2018.

\bibitem{Fischer}
Thomas~G Fischer.
\newblock Reinforcement learning in financial markets-a survey.
\newblock Report, FAU Discussion Papers in Economics, 2018.

\bibitem{Yu}
Chao Yu, Jiming Liu, and Shamim 
\newblock Reinforcement learning in healthcare: a survey.
\newblock 2019.

\bibitem{elthakeb2018releq}
Ahmed~T Elthakeb, Prannoy Pilligundla, FatemehSadat Mireshghallah, Amir
  Yazdanbakhsh, Sicun Gao, and Hadi Esmaeilzadeh.
\newblock Releq: an automatic reinforcement learning approach for deep
  quantization of neural networks.
\newblock {\em arXiv preprint arXiv:1811.01704}, 2018.

\bibitem{Ault}
James Ault, Josiah Hanna, and Guni Sharon.
\newblock Learning an interpretable traffic signal control policy.
\newblock 2019.

\bibitem{sutton2018reinforcement}
Richard~S Sutton and Andrew~G Barto.
\newblock {\em Reinforcement learning: An introduction}.
\newblock MIT press, 2018.

\bibitem{garcia2015comprehensive}
Javier Garc{\i}a and Fernando Fern{\'a}ndez.
\newblock A comprehensive survey on safe reinforcement learning.
\newblock {\em Journal of Machine Learning Research}, 16(1):1437--1480, 2015.

\bibitem{Gu}
Shixiang Gu, Timothy Lillicrap, Ilya Sutskever, and Sergey Levine.
\newblock Continuous deep q-learning with model-based acceleration.
\newblock In {\em International Conference on Machine Learning}, pages
  2829--2838.

\bibitem{Buckman}
Jacob Buckman, Danijar Hafner, George Tucker, Eugene Brevdo, and Honglak Lee.
\newblock Sample-efficient reinforcement learning with stochastic ensemble
  value expansion.
\newblock In {\em Advances in Neural Information Processing Systems}, pages
  8224--8234.

\bibitem{Feinberg}
Vladimir Feinberg, Alvin Wan, Ion Stoica, Michael~I Jordan, Joseph~E Gonzalez,
  and Sergey 
\newblock Model-based value estimation for efficient model-free reinforcement
  learning.
\newblock 2018.

\bibitem{Hausknecht}
Matthew Hausknecht and Peter Stone.
\newblock Deep recurrent q-learning for partially observable mdps.
\newblock In {\em 2015 AAAI Fall Symposium Series}.

\bibitem{van2016deep}
Hado van Hasselt, Arthur Guez, and David Silver.
\newblock Deep reinforcement learning with double q-learning.
\newblock In {\em Thirtieth AAAI conference on artificial intelligence}, 2016.

\bibitem{SchulmanTrpo}
John Schulman, Sergey Levine, Pieter Abbeel, Michael Jordan, and Philipp
  Moritz.
\newblock Trust region policy optimization.
\newblock In {\em International conference on machine learning}, pages
  1889--1897.

\bibitem{SchulmanPPO}
John Schulman, Filip Wolski, Prafulla Dhariwal, Alec Radford, and Oleg 
  preprint~arXiv:.06347 Klimov.
\newblock Proximal policy optimization algorithms.
\newblock 2017.

\bibitem{MnihAsynchronous}
Volodymyr Mnih, Adria~Puigdomenech Badia, Mehdi Mirza, Alex Graves, Timothy
  Lillicrap, Tim Harley, David Silver, and Koray Kavukcuoglu.
\newblock Asynchronous methods for deep reinforcement learning.
\newblock In {\em International conference on machine learning}, pages
  1928--1937.

\bibitem{Haarnoja}
Tuomas Haarnoja, Aurick Zhou, Pieter Abbeel, and Sergey 
  preprint~arXiv:.01290 Levine.
\newblock Soft actor-critic: Off-policy maximum entropy deep reinforcement
  learning with a stochastic actor.
\newblock 2018.

\bibitem{Lillicrap}
Timothy~P Lillicrap, Jonathan~J Hunt, Alexander Pritzel, Nicolas Heess, Tom
  Erez, Yuval Tassa, David Silver, and Daan 
  Wierstra.
\newblock Continuous control with deep reinforcement learning.
\newblock 2015.

\bibitem{Fujimoto}
Scott Fujimoto, Herke Van~Hoof, and David 
\newblock Addressing function approximation error in actor-critic methods.
\newblock 2018.

\bibitem{Wang}
Tingwu Wang, Xuchan Bao, Ignasi Clavera, Jerrick Hoang, Yeming Wen, Eric
  Langlois, Shunshi Zhang, Guodong Zhang, Pieter Abbeel, and Jimmy 
  preprint~arXiv:.02057 Ba.
\newblock Benchmarking model-based reinforcement learning.
\newblock 2019.

\bibitem{SuttonDyna}
Richard~S Sutton.
\newblock {\em Integrated architectures for learning, planning, and reacting
  based on approximating dynamic programming}, pages 216--224.
\newblock Elsevier, 1990.

\bibitem{Lampe}
Thomas Lampe and Martin Riedmiller.
\newblock Approximate model-assisted neural fitted q-iteration.
\newblock In {\em 2014 International Joint Conference on Neural Networks
  (IJCNN)}, pages 2698--2704. IEEE.

\bibitem{BartoCartpole}
Andrew~G Barto, Richard~S Sutton, man Anderson, Charles W 
  on~systems, and cybernetics.
\newblock Neuronlike adaptive elements that can solve difficult learning
  control problems.
\newblock (5):834--846, 1983.

\bibitem{Janner}
Michael Janner, Justin Fu, Marvin Zhang, and Sergey Levine.
\newblock When to trust your model: Model-based policy optimization.
\newblock In {\em Advances in Neural Information Processing Systems}, pages
  12498--12509.

\bibitem{Behzadian}
Bahram Behzadian, Soheil Gharatappeh, and Marek Petrik.
\newblock Fast feature selection for linear value function approximation.
\newblock In {\em Proceedings of the International Conference on Automated
  Planning and Scheduling}, volume~29, pages 601--609.

\bibitem{SuttonTemporal}
Richard~S Sutton.
\newblock Temporal credit assignment in reinforcement learning.
\newblock 1985.

\bibitem{Brockman}
Greg Brockman, Vicki Cheung, Ludwig Pettersson, Jonas Schneider, John Schulman,
  Jie Tang, and Wojciech 
\newblock Openai gym.
\newblock 2016.

\bibitem{dhariwal2017openai}
Prafulla Dhariwal, Christopher Hesse, Oleg Klimov, Alex Nichol, Matthias
  Plappert, Alec Radford, John Schulman, Szymon Sidor, Yuhuai Wu, and Peter
  Zhokhov.
\newblock Openai baselines, 2017.

\end{thebibliography}

\end{document}